\documentclass[letterpaper, 10 pt, conference]{ieeeconf}  
\IEEEoverridecommandlockouts

\usepackage{cite}
\usepackage{amsmath,amssymb,amsfonts}
\usepackage{algorithmic}
\usepackage{graphicx}
\usepackage{textcomp}
\usepackage{algorithm}
\usepackage{algorithmic}
\usepackage{algorithmic}
\usepackage{graphicx}
\usepackage{gensymb}
\usepackage{textcomp}
\usepackage{xcolor}
\definecolor{light_blue}{HTML}{D6EAF8}   
\usepackage{stfloats}
\usepackage{array}
\usepackage{arydshln}
\usepackage{tabularx}
\usepackage{multirow}
\usepackage{float}
\usepackage{pgfplots}
\usepackage{mwe}
\usepackage{booktabs} 
\usepackage{subfig}
\usepackage{tcolorbox}
\usepackage{tikz}
\usetikzlibrary{arrows.meta,positioning,shapes.geometric}
\usepackage{hyperref}

\title{\LARGE \bf
M2HRI: An LLM-Driven Multimodal Multi-Agent Framework for Personalized Human-Robot Interaction
}

\author{Shaid Hasan, Breenice Lee,  Sujan Sarker, and Tariq Iqbal
\thanks{The authors are with the University of Virginia, VA, USA.
        {\tt\small \{qmz9mg, pkb5ne, zzr2hs, tiqbal\}@virginia.edu}}%
}

\begin{document}

\maketitle
\thispagestyle{empty}
\pagestyle{empty}

\begin{abstract}
Multi-robot systems hold significant promise for social environments such as homes and hospitals, yet existing multi-robot systems often treat robots as functionally interchangeable, overlooking how distinct agent identities shape user perception and how such individuality changes the coordination requirements of multi-robot interaction. To address this, we introduce M2HRI, a multimodal multi-agent framework that models each robot as an identity-bearing agent through personality and long-term memory, together with a contextualized coordination mechanism that regulates agent participation. In a controlled user study ($n = 105$) in a multi-agent human–robot interaction (HRI) scenario, we found that most personality contrasts were distinguishable and consistently expressed. Long-term memory improved preference awareness and interaction naturalness, while contextualized coordination improved conversational flow, response appropriateness, and overlap avoidance. Together, these findings show that agent individuality and contextualized participation coordination play complementary roles in supporting coherent and socially appropriate multi-agent HRI. Project website available at \href{https://project-m2hri.github.io/}{https://project-m2hri.github.io/}.

\end{abstract}

\section{Introduction}
Human-Robot Interaction (HRI) has advanced considerably over the past decades, driven by the deployment of socially capable robots in everyday environments such as homes, hospitals, and educational settings \cite{lupetti2016designing}. Traditionally, HRI has focused on single-agent scenarios, where one robot interacts with a human to enable intuitive communication, effective task execution, and user engagement \cite{goodrich2008human}. These early systems established fundamental interaction mechanisms, including turn-taking, intent recognition, affective response, and anticipation \cite{iqbal2021temporal, yasar2022robots} demonstrating that robots could function as effective social partners, particularly in structured environments.

However, as real-world applications have grown in complexity and scale, the limitations of single-agent HRI have become evident. A single robot is often insufficient to simultaneously attend to multiple users, operate across shared spaces, or support the diversity of roles required in more complex social environments \cite{sheridan2016human}. This has led to a growing interest in multi-agent HRI, where multiple robots interact with a single human or group of users. In such settings, interaction is no longer centered on an individual agent, but instead emerges from the collective behavior of a team of robots. Consequently, the problem shifts from designing isolated interactions to orchestrating coherent group behavior. This transition introduces new challenges, including managing conversational turn-taking, assigning roles among agents, and maintaining a consistent and socially appropriate group identity \cite{ nigro2025social}.

\begin{figure}[!t]
    \centering
    \includegraphics[width=0.45\textwidth]
    {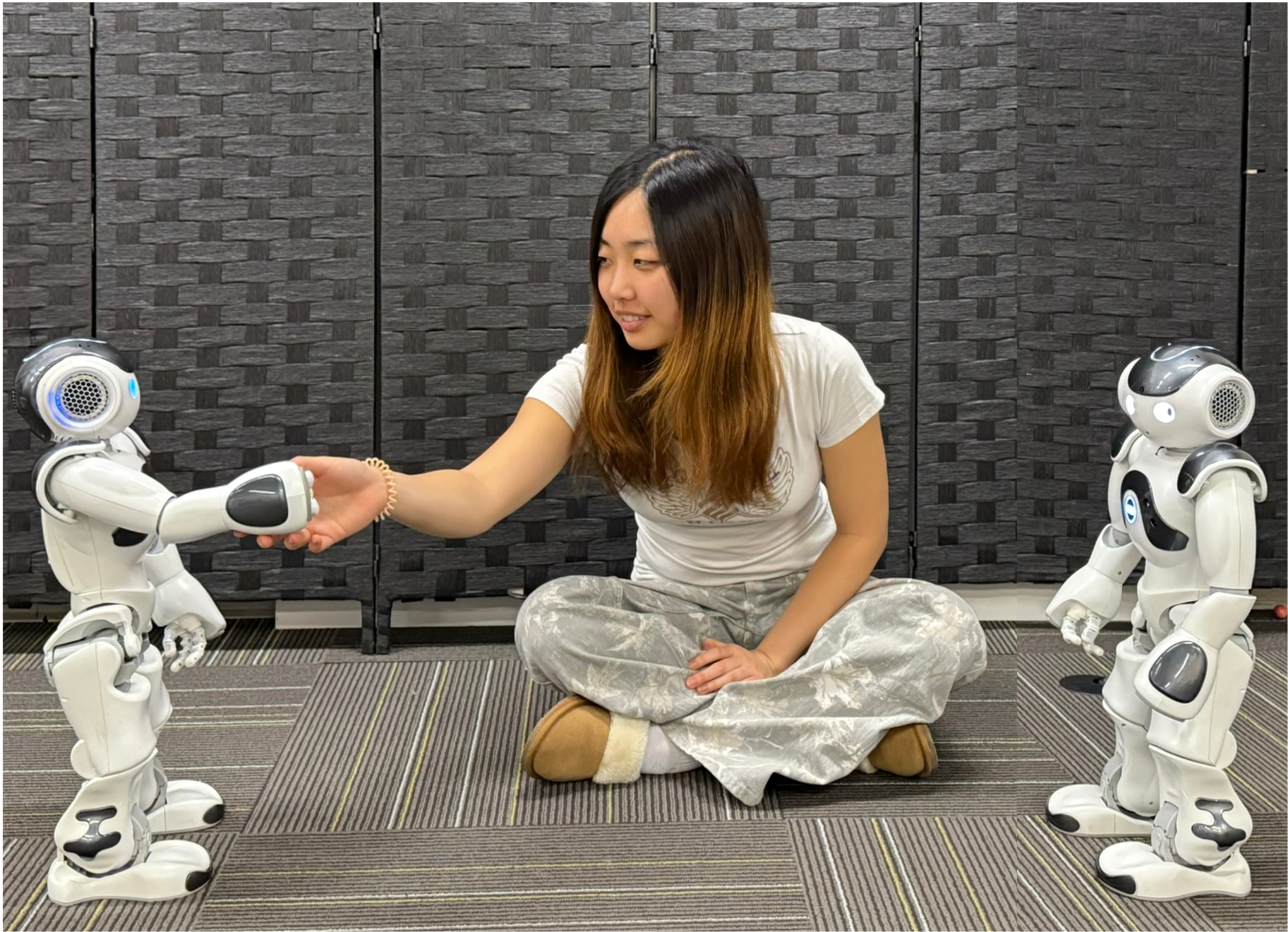}
    \caption{\small Multimodal multi-agent human-robot interaction scenario. A human user interacts with two NAO robots, each with distinct personality, memory, and perception, enabling personalized and context-aware embodied interaction.}
    \vspace{-2.0em}
    \label{fig:interaction_demo}
\end{figure}

Recent advances in vision-language models (VLMs) and large language models (LLMs) have enabled modern HRI systems to leverage multimodal perception, allowing robots to jointly reason over speech and visual observations to generate contextually appropriate responses \cite{islam2023patron,hasan2024m2rl}. However, multimodal multi-agent HRI introduces significant challenges in determining when, why, and which robot should participate in an interaction. Prior work has explored centralized coordination strategies using shared context and perceptual cues to determine turn-taking and response selection \cite{zhang2016optimal}. Research on robot-robot dynamics shows that how robots collaborate and exchange information shapes user perception, trust, and engagement \cite{erel2022carryover, green2025examining} while other work shows that robot motion, spatial behavior, embodiment, and collective dynamics shape user attention, cognition, and emotional response \cite{yasar2024imprint, yasar2021improving, yasar2024posetron,luo2024impact,yasar2023vader}. Multi-robot systems have also been investigated in application-driven contexts such as collaborative creativity and complex workflows, demonstrating both their potential and the coordination challenges they introduce \cite{pu2025beatbots,gollob2025envisioning}.

Despite this progress, a critical gap remains in how agent individual identity is modeled and maintained in multi-robot HRI. Existing multi-robot HRI works treat robots as functionally interchangeable, overlooking differences in personality and interaction history, even though appropriate behavior in shared social settings depends on the individual. A robot that maintains a consistent personality and remembers user preferences creates a different experience from one that treats every interaction as new, especially in multi-robot settings where individuality helps users perceive the robots as distinct social agents rather than identical machines operating side by side. Introducing agent individuality also changes the coordination problem. Determining which robot should respond can no longer rely only on generic turn-taking rules; participation must be selected according to the interaction context and the characteristics of each agent. To address this gap, we investigate three research questions: 

\begin{itemize}
    \item \textbf{RQ1:} How does robot personality shape users’ perceptions of individual agents in multi-agent HRI?
    \item \textbf{RQ2:} How does agent-specific long-term memory influence personalization and contextual awareness in multi-robot interaction?
    \item \textbf{RQ3:} How does context-aware coordination among agents affect interaction quality?
\end{itemize}

To address these research questions, we design and implement M2HRI, a multimodal multi-agent HRI framework built on LLMs and VLMs that integrates multimodal perception with identity-bearing agents and contextualized participation coordination within a unified system (Fig. \ref{fig:interaction_demo}). We evaluate the framework through a controlled human-subject study ($n = 105$), varying agent personality, memory, and coordination. Our results show that personality enables users to reliably distinguish between agents, recognize intended traits, and enhances interaction engagement, though its effectiveness varies with how behaviorally expressive a trait is. Memory significantly improves recall accuracy, preference awareness, and perceived personalization, while coordination plays a critical role in ensuring coherent interaction by improving conversational flow, response appropriateness, and reducing overlap. Together, these findings indicate that modeling agent identity and coordinating the participation of non-interchangeable agents are complementary design requirements for coherent, personalized, and socially appropriate multi-robot interaction.
\section{Related Work}
\subsection{Multi-Agent HRI}
Research in multi-agent HRI shows that interacting with multiple robots introduces a fundamentally different and more complex social context than single-robot settings, shaping human cognition, behavior, and emotion \cite{saadon2025scammed, luo2024impact}. Prior work demonstrates that the mere presence, number, and coordination patterns of multiple robots can significantly influence task performance, feelings of being observed, and emotional responses, highlighting the role of embodiment and collective motion in interaction outcomes \cite{luo2024impact}. Beyond presence effects, multi-robot interactions give rise to rich social dynamics such as conformity, compliance, and group perception, where humans adapt their behavior based on robot group behavior and structure \cite{saadon2025scammed,wright2025know}. In particular, studies on robot–robot relationships show that ingroup and outgroup dynamics among robots influence human compliance, trust, and collaboration, altering how people respond to instructions and perceive robotic teams \cite{wright2025know}. At the system level, recent work highlights emerging challenges in orchestration, coordination, and interaction management in multi-agent systems, while practical HRI studies reveal difficulties in controlling and designing interactions involving multiple robots \cite{schombs2025conversation, bejarano2024hardships}. However, existing approaches primarily address behavioral or system-level coordination while giving limited attention to how participation should be managed when robots maintain distinct identities and interaction histories.

\subsection{Robot Personality and Memory in HRI}
Robot personality has been widely studied in single-agent HRI, with the Big Five Factor Model serving as the dominant framework for operationalizing distinct traits through speech, gesture, and language style, and studies  finding that personality traits influence trust and engagement, and that individual user traits moderate these effects \cite{lim2022we, nardelli2024bpersonality}. Research has focused heavily on extraversion while beginning to explore other dimensions such as neuroticism and conscientiousness, with recent LLM-driven approaches enabling more scalable and multi-trait personality expression \cite{zhang2025exploring,nardelli2024bpersonality }. In parallel, work on robot memory demonstrates that agents capable of storing and recalling user preferences across sessions are perceived as more personalized, contextually aware, and relationally engaging \cite{ligthart2022memory,paplu2022harnessing}. Despite these advances, personality, memory, and continual representation learning have largely been examined as separate capabilities in single-agent settings \cite{yasar2023coral}. How these complementary components make robots appear as distinct individuals, and how such individuality affects multi-agent coordination, remains underexplored.
\section{M2HRI Framework}
M2HRI is a co-located multimodal multi-agent human–robot interaction framework (Figure \ref{fig:m2hri}) that enables natural, coordinated interaction between a human user and a team of robot agents. Built on large language models (LLMs) and vision-language models (VLMs), the framework is designed around two complementary goals: maintaining the individuality of each agent through personality and long-term memory, and coordinating the participation of these non-interchangeable agents to produce coherent group behavior. The following subsections describe the agent architecture, contextualized coordination mechanism, and system implementation.

\subsection{Agent Architecture}
\begin{figure*}[!t]
\centering
  \includegraphics[width=0.93\textwidth]{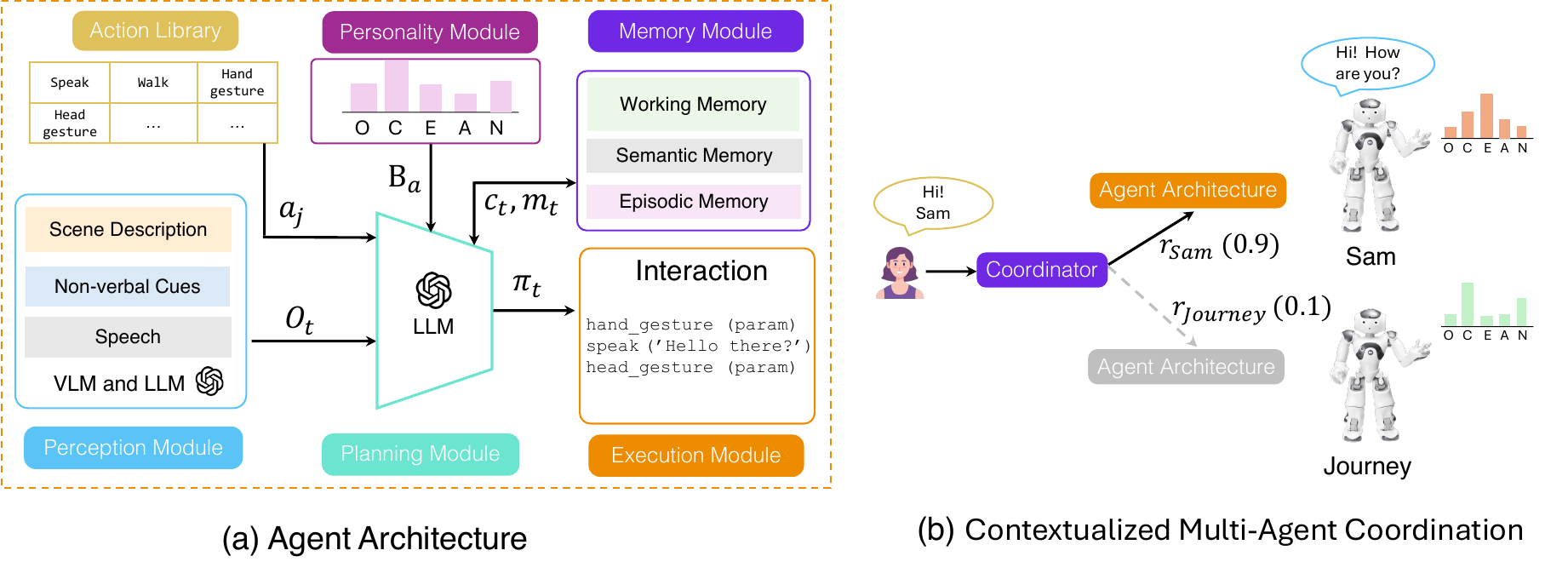}
  \caption{\small M2HRI framework. (a) Agent architecture showing perception, personality, memory, planning, and action modules. (b) Human multi-agent interaction with contextualized multi-agent coordination, implemented through a centralized coordinator.}
  \vspace{-1em}
  \label{fig:m2hri}
\end{figure*}

We model each robot agent $a_i$ as, $a_i = \langle \phi_i, \Lambda_i, \Gamma_i \rangle$ (Figure \ref{fig:m2hri}a).
where $\phi_i$ denotes the \textit{agent identity profile}, $\Gamma_i$ its action capability set and $\Lambda_i$ denotes the \textit{cognitive loop}. This decomposition separates \textit{what an agent is} from \textit{how it operates}. The \textit{agent identity profile} is defined as:
\vspace{-0.5em}
\begin{equation}
\phi_i = \langle M_i, B_i \rangle
\label{eq:profile}
\vspace{-0.5em}
\end{equation}
where $M_i$ denotes the agent’s memory, and $B_i$ its personality representation. The \textit{cognitive loop} is defined as: $\Lambda_i = \langle P_i, R_i, E_i \rangle$,
where $P_i$ denotes the perception module, $R_i$ the reasoning (planning) module, and $E_i$ the execution module. Together, these form a continuous perception-cognition-action loop that transforms multimodal input into embodied robot behavior. While all agents share the same underlying architecture, their individual profiles $\phi_i$ differ: each agent maintains its own personality $B_i$, and memory $M_i$. As a result, behavior is conditioned on these agent-specific factors, leading to differentiated responses despite a common framework. The following subsections describe each component in detail.

\subsubsection{Perception}
The Perception module $P_i$ is the entry point of each agent's cognitive loop, transforming raw multimodal sensory input into a semantic observation that grounds all downstream memory retrieval and planning. At interaction timestep $t$, agent $a_i$ receives a multimodal input vector, $\mathbf{x}_t = \left\langle x_t^{\text{speech}},\, x_t^{\text{vision}} \right\rangle$, where $x_t^{\text{speech}}$ denotes the spoken utterance transcribed from the interaction and $x_t^{\text{vision}}$ denotes the visual frame captured by the robot agent's onboard camera. The perception function maps these modalities to a unified semantic observation:
\vspace{-0.5em}
\begin{equation}
    o_t = P_i\!\left(x_t^{\text{speech}},\, x_t^{\text{vision}}\right)
    \label{eq:observation}
    \vspace{-0.5em}
\end{equation}
$P_i$ is implemented using a vision language model (VLM) that jointly processes speech and visual input to produce a coherent natural-language description of the current interaction context. This includes grounding spoken references in the physical scene, identifying entities and their spatial relationships, and inferring non-verbal cues such as user gaze direction or gesture. The resulting observation $o_t$ is a textual representation that can be directly consumed by the downstream Memory and Planning modules. 

\subsubsection{Memory}

The Memory Module provides each agent with persistent and contextual information processing capabilities, enabling both long-term knowledge retention and short-term context awareness. The memory module is organized into three distinct submodules:
\begin{itemize}
    \item \textbf{Working memory}, which maintains a short-term interaction context consisting of recent conversational history and the current interaction state. This memory supports real-time reasoning and decision-making by providing an active contextual buffer for ongoing cognition \cite{baddeley2020working}.
    
    \item \textbf{Semantic memory}, which stores factual knowledge about users, objects, and the environment (e.g., identities, attributes, and relationships), enabling consistent understanding across interactions \cite{binder2011neurobiology}.
    
    \item \textbf{Episodic memory}, which captures temporally grounded interaction experiences, including user preferences, past behaviors, and situational context, allowing agents to adapt based on prior interactions \cite{tulving2002episodic}.
\end{itemize}

Each agent $a_i$ maintains an agent-specific long-term memory store $M_{a_i}$, ensuring distinct interaction histories across agents. Given an observation $o_t$, the agent retrieves relevant memories as $m_t = R_{\text{mem}}(o_t, M_{a_i})$ and updates its memory with newly acquired information as $M_{a_i} \leftarrow M_{a_i} \cup S_{\text{mem}}(o_t, M_{a_i})$. Short-term context is maintained through a working memory $c_t = \{o_{t-k}, \ldots, o_{t-1}, o_t\}$, a sliding window of recent observations that is reset at the start of each session and not persisted to disk. An LLM-based controller performs context-sensitive retrieval and storage while filtering redundant content.

\subsubsection{Personality}
The \textit{Personality Module} shapes each agent’s reasoning and interaction behavior through an explicit and persistent personality representation. Rather than acting as a stylistic layer, personality is integrated directly into the agent’s cognitive process, influencing how it formulates responses, engages in conversation, and selects actions. Each agent's personality is modeled using the Five-Factor Model (FFM) \cite{barrick1991big}, also known as the Big Five and represented as a five-dimensional vector:
\vspace{-0.5em}
\begin{equation}
\mathbf{B}_{a_i} = (O, C, E, A, N)
\label{eq:ocean}
\vspace{-0.5em}
\end{equation}
corresponding to Openness to Experience (O), Conscientiousness (C), Extraversion (E), Agreeableness (A), and Neuroticism (N). Each dimension is discretized on a scale from 1 (low) to 5 (high), with 3 as a neutral midpoint, enabling interpretable and controllable variation across agents. To interface with language models, trait values are mapped to natural-language descriptors using a deterministic function $f_{\text{trait}} : \{1,2,3,4,5\} \rightarrow \text{linguistic modifiers}$. Extreme values yield intensified descriptors (e.g., ``extremely outgoing''), intermediate values produce softer modifiers (e.g., ``somewhat cautious''), and the neutral value results in no explicit descriptor. These descriptors are injected into the agent’s system, allowing the planner to produce behavior that remains consistent with the agent’s intended disposition.

\subsubsection{Interaction Planner}
This module serves as the central reasoning component of each agent, integrating current perception, memory, personality, interaction context, and available action to generate embodied interaction policies. M2HRI adopts a language-model-driven approach in which reasoning and action selection are jointly performed by an LLM conditioned on these inputs. At timestep $t$, the planner takes the current observation $o_t$, interaction context $c_t$, retrieved memory $m_t$, personality vector $B_{a_i}$, and action capability library $\mathcal{A}$ to produce an embodied interaction policy: $\pi_t = \text{Planner}(o_t, c_t, m_t, B_{a_i},\mathcal{A} )$.
The policy $\pi_t$ is represented as an ordered list of parameterized actions:
\vspace{-0.5em}
\begin{equation}
\pi_t = \{(a_1, \theta_1), (a_2, \theta_2), \dots, (a_k, \theta_k)\} 
\quad a_j \in A
\label{eq:param_policy}
\vspace{-0.5em}
\end{equation}
where each $a_j \in \mathcal{A}$ is an executable primitive (e.g., speech, gesture, locomotion) and $\theta_j$ its associated parameters. Depending on the context, the policy may consist of a single action or a multi-step sequence.

\subsubsection{Execution}
The Execution Module $E$ is the runtime component that executes planned policies as embodied behavior. Given a parameterized policy $\pi_t$ (Eq.~\ref{eq:param_policy}) generated by the Interaction Planner, the execution module maps it to executable behavior:
\vspace{-0.5em}
\begin{equation}
u_t = E(\pi_t), \qquad u_t \in \Gamma_i
\vspace{-0.5em}
\end{equation}
where $u_t$ represents the embodied behavior produced at timestep $t$, constrained to the agent's capability set $\Gamma_i$. M2HRI represents behavior using a finite set of reusable action primitives, including speech, gestures, posture changes, head movements, and locomotion. Each primitive is parameterized to specify its content and style, such as spoken text, gesture type, or motion direction. Execution proceeds sequentially over the policy, enabling compound behaviors such as ``gesture $\rightarrow$ speak $\rightarrow$ nod.''

\subsection{Contextualized Multi-Agent Coordination}
M2HRI adopts a contextualized coordination mechanism (Figure \ref{fig:m2hri}b) to regulate interaction among multiple agents while preserving decentralized cognition. Each agent independently maintains its perception-planning-execution loop, while a shared coordinator governs interaction turn-taking and agent participation. At timestep $t$, each agent $a_i$ produces a local observation $o_t^{(i)}$. The coordinator aggregates these observations together with the interaction context $c_t$ and each agent's personality vector $B_{a_i}$ into a collective set:
\vspace{-0.5em}
\begin{equation}
    \mathcal{Z}_t = \left\{z_t^{(i)}\right\}_{i=1}^{N}, 
    \quad z_t^{(i)} = \left(o_t^{(i)},\, c_t,\, B_{a_i}\right)
    \vspace{-0.5em}
\end{equation}
The coordinator then jointly reasons over $\mathcal{Z}_t$ to assign a participation-suitability score to each agent:
\vspace{-0.5em}
\begin{equation}
    r_{i,t} = f_{\text{coord}}\!\left(\mathcal{Z}_t\right)_i, 
    \quad r_{i,t} \in [0,1]
    \vspace{-0.5em}
\end{equation}
where $f_{\text{coord}}(\cdot)$ is an LLM-based coordination function that evaluates how appropriate it is for agent $a_i$ to participate given its observation, personality, and shared context. Agents whose scores exceed a predefined threshold $\tau$ are selected to participate: $S_t = \{\, i \mid r_{i,t} \ge \tau \,\}$.
Selected agents respond sequentially, each independently generating and executing its own embodied policy.

\subsection{System Implementation} 
M2HRI is deployed on two NAO humanoid robots with onboard RGB cameras, microphones, and speakers, while cognitive processing runs on an external LLM workstation. Perception is implemented through continuous camera frame acquisition, where frames are base64-encoded and passed to a VLM for scene grounding. Memory is implemented using two independent \texttt{ChromaDB} collections (semantic, episodic), with entries embedded via \texttt{text-embedding-3-large} and access mediated by an LLM-driven \texttt{LangGraph} memory agent that handles retrieval and storage at runtime. Personality is parameterized as integer Big Five values mapped to natural-language descriptors in each agent’s system prompt. Planning is implemented with \texttt{AutoGen} AssistantAgents backed by \texttt{GPT-4o-mini}, constrained to produce structured tool calls from a dynamically generated JSON schema, which are serialized over \texttt{ZMQ} and executed on the NAO client via \texttt{NaoQi} proxies. Coordination is implemented by a centralized LLM-based coordinator that evaluates each agent's suitability to participate based on visual input, dialogue context, and personality descriptors, producing participation-suitability scores.

\section{User Study}
\subsection{Study Design} We conducted a controlled observer study to assess the three research questions. Participants (n = 105) were assigned to one of seven experimental blocks, with 15 participants in each block. The five personality blocks used separate participant groups, whereas the memory and coordination blocks used paired within-participant comparisons. Each session lasted approximately 10 minutes and involved watching pre-recorded interaction videos followed by a questionnaire. We adopted a video-based observer approach to enable strict experimental control by minimizing confounds from robot motion, environmental conditions, and interaction timing present in live deployments. Using pre-recorded videos ensures all participants observe identical interactions, so differences in responses can be attributed directly to the manipulated variables. The study protocol was approved by the Institutional Review Board (IRB).

\subsection{Participants}
We recruited 105 participants via \textit{Prolific} platform. Participants ranged in age from 18 to 77 years ($M=39.3, SD=13.7$) and included 66 males, 38 females, and 1 non-binary participant. Education levels included college degree (34.3\%), some college (27.6\%), high school degree (18.1\%), graduate degree (17.1\%), and some graduate education (2.9\%). Prior robot experience was low to moderate on a 5-point Likert scale: 1 (37.1\%), 2 (35.2\%), 3 (21.0\%), 4 (4.8\%), and 5 (1.9\%).

\begin{figure*}[!t]
\centering
  \includegraphics[width=0.95\textwidth]{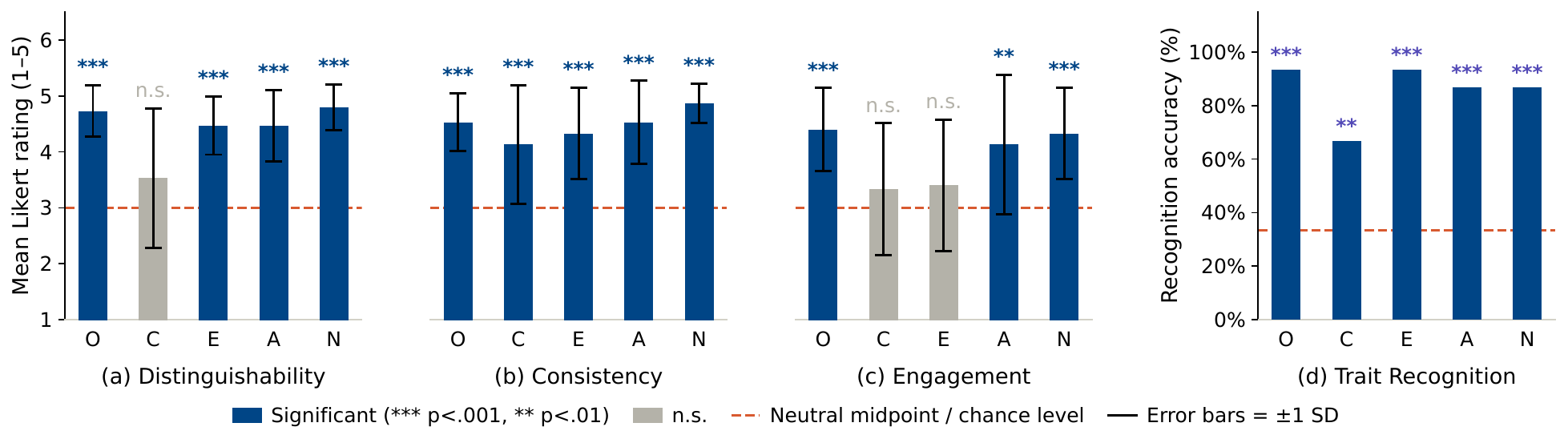}
      \vspace{-0.5em}
  \caption{\small Personality evaluation results (RQ1). Mean Likert ratings for (a) distinguishability, (b) consistency, and (c) engagement across five Big Five trait conditions (O = Openness, C = Conscientiousness, E = Extraversion, A = Agreeableness, N = Neuroticism). The dashed line indicates the neutral midpoint ($\mu_0 = 3.0$) (d) Personality Trait recognition accuracy; the dashed line indicates chance level (33.3\%).
}
  \vspace{-1em}
  \label{fig:personality_result}
\end{figure*}

\subsection{Experimental Conditions}
We designed seven experimental blocks across three dimensions: five personality blocks, one paired memory block, and one paired coordination block. Each variant was recorded as a stimulus video of a human interacting with two NAO robots.

\smallskip
\noindent\textbf{Personality Conditions.}
To assess whether users can perceive and differentiate agent personalities, we designed five personality conditions corresponding to each dimension of the Big Five model (O, C, E, A, N). In each condition, one robot was configured at the high extreme of the target trait (trait value $= 5$) and the other at the low extreme (trait value $= 1$), while all remaining trait dimensions were held constant at the neutral midpoint (trait value $= 3$) for both the robot agents. For each condition, we selected an interaction topic designed to naturally highlight observable behavioral differences between the two agents along the target dimension. Specifically, Openness was evaluated through a discussion of \textit{``unconventional hobbies"}; Conscientiousness through \textit{``choosing between studying for an exam vs. attending a party''}; Extraversion through \textit{``meeting new people"}; Agreeableness through \textit{``providing feedback on a podcast idea"}; and Neuroticism through \textit{``reacting to losing a pet"}.

\smallskip
\noindent\textbf{Memory Condition.}
To assess whether memory enables more personalized and contextually aware interactions, we compared a \textit{with-memory} and a \textit{without-memory} variant using a consistent interaction scenario \textit{``recalling and using user’s shared preferences (e.g., favorite color, food, activities)"}. In the \textit{with-memory} condition, each agent operated with both working memory and its full long-term memory module active, allowing it to maintain short-term context while storing and retrieving semantic and episodic information. In contrast, in the \textit{without-memory} condition, long-term memory was disabled, and agents relied solely on working memory. Across both conditions, the interaction topic, agent personality configurations, and user script were kept identical.

\smallskip
\noindent\textbf{Coordination Condition.}
To assess the effect of contextualized participation on interaction we compared a \textit{with-coordination} and a \textit{without-coordination} variant using a consistent interaction scenario \textit{``discussion on AI in education with direct addressing of individual robots"}. In the \textit{with-coordination} condition, we used M2HRI’s contextualized coordination mechanism, which computes per-agent response likelihood scores based on each agent’s local observation, personality, and conversation context, and selects which agent(s) should respond. In the \textit{without-coordination} condition, the contextualized coordinator was removed, and each robot independently made its own turn-taking decisions using only local context. Across both conditions, the interaction topic and personality configurations were kept identical.

\section{Results and Analysis}
We analyze the results corresponding to the three research questions: personality (RQ1), memory (RQ2), and coordination (RQ3). For personality, we use one-sample t-tests to assess deviations from the neutral midpoint, along with Wilcoxon signed-rank tests. For memory and coordination, we use paired-sample t-tests to compare conditions, along with Wilcoxon signed-rank tests. We report both parametric and non-parametric tests to ensure robustness, given the ordinal nature of Likert-scale data.

\begin{figure*}[!t]
\centering
  \includegraphics[width=0.88\textwidth]{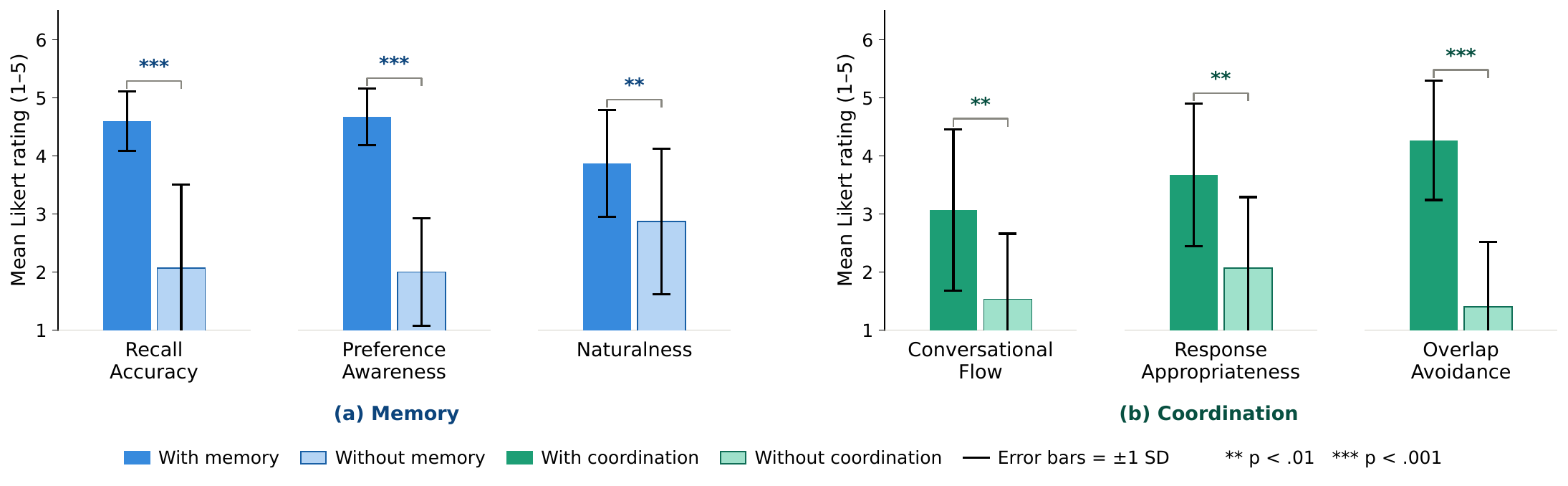}
    \vspace{-0.5em}
  \caption{\small Memory (RQ2) and coordination (RQ3) evaluation results. Paired bar charts compare with- and without-condition means across three measures each. (a) Memory measures: recall accuracy, preference awareness, naturalness. (b) Coordination measures: conversational flow, response appropriateness, overlap avoidance. Brackets indicate statistical significance with ** $p < .01$, *** $p < .001$.
}
  \vspace{-1em}
  \label{fig:mem_coord}
\end{figure*}

\subsection{Personality (RQ1)}
\subsubsection{Distinguishability of Robot Personalities}
We first evaluated whether participants perceived the two robots as having clearly distinguishable personalities across all five personality trait conditions (Fig. \ref{fig:personality_result}a). Across conditions, mean ratings for personality distinguishability were consistently above the neutral midpoint (3), with particularly strong effects observed for Openness ($M = 4.73, SD = 0.46$), Neuroticism ($M = 4.80, SD = 0.41$), Agreeableness ($M = 4.47, SD = 0.64$), and Extraversion ($M = 4.47, SD = 0.52$). Conscientiousness showed comparatively lower ratings ($M = 3.53, SD = 1.25$), closer to neutrality. One-sample t-tests against the neutral midpoint confirmed that distinguishability was significantly above the neutral midpoint for four of the five traits ($p < .001$), while Conscientiousness did not reach statistical significance ($p = .120$). Wilcoxon signed-rank tests showed a consistent pattern, with significant deviations from the neutral median for all traits except Conscientiousness. Effect sizes were large for Openness ($d = 3.79$), Neuroticism ($d = 4.35$), Agreeableness ($d = 2.29$), and Extraversion ($d = 2.84$), indicating strong perceptual differences. In contrast, Conscientiousness yielded a smaller effect ($d = 0.43$).

\noindent \textbf{Summary:}
Participants were generally able to clearly distinguish between robot personalities, suggesting that the personality representation produced perceptible differences between agents. However, this effect varies by trait: Openness and Neuroticism produced strong perceptual separation, whereas Conscientiousness was less salient. This suggests that not all personality dimensions are equally observable in short interaction scenarios, and that distinguishability depends on how behaviorally expressive a trait is.

\vspace{0.5em}

\subsubsection{Consistency of Personality Expression}
We next assessed whether each robot maintained a consistent personality throughout the interaction (Fig. \ref{fig:personality_result}b). Across all five traits, ratings of personality consistency were significantly above the neutral midpoint. Mean ratings were high for Neuroticism ($M = 4.87, SD = 0.35$), Openness ($M = 4.53, SD = 0.52$), Agreeableness ($M = 4.53, SD = 0.74$), Extraversion ($M = 4.33, SD = 0.82$), and Conscientiousness ($M = 4.13, SD = 1.06$). One-sample t-tests confirmed statistical significance for all five traits (all $p < .001$), indicating that participants consistently perceived stable personality expression across conditions. Wilcoxon signed-rank tests further supported these findings, with all traits showing significant deviations from the neutral median. Effect sizes were large across all traits, including Neuroticism ($d = 5.31$), Openness ($d = 2.97$), Agreeableness ($d = 2.06$), Extraversion ($d = 1.63$), and Conscientiousness ($d = 1.07$).

\noindent \textbf{Summary:}
Personality was not only perceptible but also consistently maintained throughout the interaction. Unlike distinguishability, which varied across traits, consistency remained robust across all conditions. This suggests that the personality representation produced behavior perceived as consistent across the observed interaction.

\vspace{0.5em}

\subsubsection{Perceived Engagement Across Personality Conditions}
We evaluated whether robot personalities contributed to a more engaging interaction (Fig. \ref{fig:personality_result}c). Engagement ratings showed greater variability across traits compared to distinguishability and consistency. High engagement was observed for Openness ($M = 4.40, SD = 0.74$) and Neuroticism ($M = 4.33, SD = 0.82$), both significantly above the neutral midpoint ($p < .001$), as well as for Agreeableness ($M = 4.13, SD = 1.25$), which was also significantly above neutral ($p = .003$). In contrast, Extraversion ($M = 3.40, SD = 1.18$) and Conscientiousness ($M = 3.33, SD = 1.18$) did not significantly differ from neutral. Effect sizes reflected this variation, with large effects for Openness ($d = 1.90$), Neuroticism ($d = 1.63$), and Agreeableness ($d = 0.91$), and small effects for Extraversion ($d = 0.34$) and Conscientiousness ($d = 0.28$).

\noindent \textbf{Summary:}
Perceived engagement varies across personality traits and its impact depends strongly on the specific trait being expressed. Traits that shape the emotional and affective tone of the interaction (e.g., Openness, Neuroticism, and Agreeableness) contribute more strongly to engagement, whereas Extraversion and Conscientiousness have a weaker effect.

\vspace{0.5em}

\subsubsection{Trait Recognition}
To assess whether participants could identify the intended personality traits, we analyzed participant selections using binomial tests against chance level ($1/3$). Participants correctly identified the high-trait robot at rates significantly above chance for all five traits (Fig. \ref{fig:personality_result}d). Accuracy was highest for Openness and Extraversion (93.3\%), followed by Agreeableness and Neuroticism (86.7\%), and lowest for Conscientiousness (66.7\%). Binomial tests confirmed statistical significance across all traits (all $p \leq .0085$), with strong confidence intervals indicating reliable detection of the intended personality differences.

\noindent \textbf{Summary:}
These results provide strong complementary evidence that participants could reliably recognize and attribute personality traits to the robots. Even for Conscientiousness, which showed weaker effects in Likert ratings, participants still identified the intended trait above chance. This suggests that personality differences are perceptible at a categorical level, even when they are less strongly reflected in subjective ratings of distinguishability or engagement.

\subsection{Memory (RQ2)}

We evaluated the effect of long-term memory on user perception by comparing \textit{with-memory} and \textit{without-memory} conditions across three measures (Fig. \ref{fig:mem_coord}a): recall accuracy, awareness of user preferences, and interaction naturalness.

Paired-sample t-tests confirmed that these differences were statistically significant across all measures. Recall accuracy showed a strong effect ($t(14) = 6.52$, $p < .001$, 95\% CI $[1.70, 3.37]$), as did awareness of user preferences ($t(14) = 12.65$, $p < .001$, 95\% CI $[2.22, 3.12]$). Interaction naturalness also showed a significant improvement ($t(14) = 3.09$, $p = .008$, 95\% CI $[0.31, 1.69]$). Wilcoxon signed-rank tests further supported these findings. All three measures were significantly higher in the with-memory condition compared to the without-memory condition: recall accuracy ($Z = -3.24$, $p = .001$), awareness of user preferences ($Z = -3.53$, $p < .001$), and interaction naturalness ($Z = -2.39$, $p = .017$). Effect size analysis indicated large effects for recall accuracy ($d = 1.68$) and awareness of user preferences ($d = 3.27$), and a moderate effect for interaction naturalness ($d = 0.80$).

\noindent \textbf{Summary:}
These results indicate that long-term memory strongly supports perceived personalized and contextually aware interaction in multi-agent settings. Memory primarily enhances the robots' ability to accurately recall prior information and adapt to user preferences, leading to strong improvements in perceived personalization. While memory also contributes to more natural interactions, its effect on naturalness is comparatively smaller, suggesting that natural conversational flow depends on additional factors beyond memory alone.

\subsection{Contextualized Coordination (RQ3)}
We evaluated the effect of contextualized participation coordination on conversational quality by comparing \textit{with-coordination} and \textit{without-coordination} conditions across three measures (Fig. \ref{fig:mem_coord}b): conversational flow, appropriateness of the responding agent, and absence of interruption or overlap.

Paired-sample t-tests confirmed that these differences were statistically significant across all measures. Conversational flow showed a significant improvement ($t(14) = 4.08$, $p = .001$, 95\% CI $[0.73, 2.34]$), as did response appropriateness ($t(14) = 4.00$, $p = .001$, 95\% CI $[0.74, 2.46]$). Overlap avoidance demonstrated the strongest effect ($t(14) = 7.89$, $p < .001$, 95\% CI $[2.09, 3.65]$). Wilcoxon signed-rank tests further supported these findings. All three measures were significantly higher in the with-coordination condition: conversational flow ($Z = -2.70$, $p = .007$), response appropriateness ($Z = -2.69$, $p = .007$), and overlap avoidance ($Z = -3.35$, $p < .001$). Effect size analysis revealed large effects across all measures. Conversational flow ($d = 1.05$) and response appropriateness ($d = 1.03$) showed large effects, while overlap avoidance exhibited a very large effect ($d = 2.04$).

\noindent \textbf{Summary:}
These results indicate that contextualized participation coordination plays a critical role in maintaining coherent multi-agent interaction. Coordination improves both who speaks and how the interaction unfolds, leading to more appropriate responses and smoother conversational flow. Its strongest impact is in preventing interruptions and overlapping behavior, which are key sources of breakdown in multi-agent dialogue. Overall, the findings highlight that effective coordination is essential for maintaining conversational structure and ensuring that multi-robot interactions are perceived as organized and natural rather than chaotic or conflicting.

\vspace{-1em}
\section{Discussion}
\subsection{Effects of Robot Personality (RQ1)}
The strong distinguishability and high trait recognition accuracy observed across most personality conditions indicate that personality, when encoded into agent behavior, produces behaviorally clear differences that users can readily perceive and categorize. This is consistent with prior findings in social cognition showing that humans form personality impressions from brief behavioral cues \cite{ambady1992thin}, especially when traits are expressed through observable language and affect. However, the perceptibility of personality encoding depends on how behaviorally expressive a trait is. Traits such as \textit{Openness} and \textit{Neuroticism} were more easily distinguishable because they appear through immediately observable cues, including differences in language style, emotional tone, and variability in responses. In contrast, \textit{Conscientiousness} was less noticeable in short interactions, as it is typically expressed through long-term, goal-directed behavior rather than conversational style \cite{john1993determinants}. Despite this difference, the consistently high ratings for personality stability indicate that M2HRI produced behavior perceived as consistent across the observed interaction, which is important for helping users recognize each agent as distinct. Engagement did not mirror distinguishability. Extraversion was clearly distinguishable but did not receive significantly above-neutral engagement ratings, suggesting that a trait’s emotional tone matters more than how noticeable it is. Overall, these findings suggest that a consistent personality helps users perceive each robot as distinct, though its clarity and effect on engagement depend on how the trait is expressed.

\subsection{Effects of Robot Memory (RQ2)}
Our results indicate that agents' memory significantly improves recall accuracy and users' preference awareness, confirming that long-term memory is a key mechanism for enabling personalization in LLM-driven robot agents. Without long-term memory, agents operate in a stateless manner, producing responses that are locally coherent but lack continuity across interactions. While LLMs can integrate retrieved context into responses, natural interaction also depends on factors such as timing, and coordination. This helps explain the comparatively smaller improvement observed in interaction naturalness. Even in the absence of long-term memory, agents still rely on working memory, which is sufficient to maintain short-term conversational coherence within a session. As a result, the without-memory condition was not chaotic or broken, but rather impersonal. This highlights an important distinction: naturalness and personalization are not the same. While naturalness can be supported by short-term context alone, long-term memory plays a more decisive role in enabling personalized, user-specific interactions.

\subsection{Effects of Contextualized Coordination (RQ3)}
The coordination results highlight that achieving coherent multi-agent interaction is fundamentally a structural challenge. In the absence of coordination, agents rely only on local context, often leading to simultaneous responses and conversational overlap. This violates basic turn-taking norms in human dialogue and is perceived as a breakdown in social behavior rather than a simple technical issue \cite{skantze2021turn}. M2HRI addresses this through a shared contextualized decision layer, implemented centrally, that evaluates which agent is most appropriate to participate. This not only prevents overlap but also improves response appropriateness, since selecting the right agent requires more than perceptual access, it requires understanding which agent is socially and contextually best suited to respond. By incorporating personality into this process, M2HRI enables more socially grounded response selection, aligning with the idea that users apply human-like expectations when evaluating robots \cite{reeves1996media}. Overall, these results show that coordination is not merely an optimization but necessary for creating a coherent and socially appropriate multi-agent system.
\section{Conclusion}
In this work, we presented M2HRI, a multimodal multi-agent HRI framework that models robots as identity-bearing agents through persistent personality and agent-specific long-term memory, together with contextualized participation coordination. Through a controlled video-based user study, we found that personality and memory helped users distinguish robot agents and perceive interactions as more personalized, while contextualized coordination improved conversational flow, response appropriateness, and overlap avoidance. These findings show that distinct agent identities, continuity across interactions, and coordinated participation are complementary requirements for coherent and personalized multi-robot interaction. Future work will include in-person, adaptive personality modeling, and long-term deployments.

\section*{ACKNOWLEDGMENTS}
Research reported in this paper is supported by the \\ National Science Foundation under Award No. 2441587.

\bibliographystyle{IEEEtran}
\bibliography{references}

\end{document}